\documentclass{article}

\usepackage{arxiv}

\usepackage[utf8]{inputenc} % allow utf-8 input
\usepackage[T1]{fontenc}    % use 8-bit T1 fonts
\usepackage{hyperref}       % hyperlinks
\usepackage{url}            % simple URL typesetting
\usepackage{booktabs}       % professional-quality tables
\usepackage{amsfonts}       % blackboard math symbols
\usepackage{nicefrac}       % compact symbols for 1/2, etc.
\usepackage{microtype}      % microtypography
\usepackage{lipsum}
\usepackage{graphicx}
\graphicspath{ {./images/} }
\usepackage{caption}
\usepackage{float}
\usepackage{amsmath}

\title{Real-time Pollutant Identification through Optical PM Micro-Sensor}

\author{
  Elie Azeraf, PhD \\
  Capgemini Engineering \\
  \texttt{elie.azeraf@capgemini.com} \\
   \And
  Audrey Wagner \\
  Capgemini Engineering \\
  \texttt{audrey.wagner@capgemini.com} \\
  \And
  Emilie Bialic, PhD \\
  Capgemini Engineering \\
  \texttt{emilie.bialic@capgemini.com} \\
   \And
  Samia Mellah, PhD \\
  Capgemini Engineering \\
  \texttt{samia.mellah@capgemini.com} \\
  \And
  Ludovic Lelandais, PhD \\
  Capgemini Engineering \\
  \texttt{ludovic.lelandais@capgemini.com} \\
}

\begin{document}
\maketitle
\begin{abstract}
Air pollution remains one of the most pressing environmental challenges of the modern era, significantly impacting human health, ecosystems, and climate. While traditional air quality monitoring systems provide critical data, their high costs and limited spatial coverage hinder effective real-time pollutant identification. Recent advancements in micro-sensor technology have improved data collection but still lack efficient methods for source identification. This paper explores the innovative application of machine learning (ML) models to classify pollutants in real-time using only data from optical micro-sensors. We propose a novel classification framework capable of distinguishing between four pollutant scenarios: Background Pollution, Ash, Sand, and Candle. Three Machine Learning (ML) approaches—XGBoost, Long Short-Term Memory networks, and Hidden Markov Chains — are evaluated for their effectiveness in sequence modeling and pollutant identification. Our results demonstrate the potential of leveraging micro-sensors and ML techniques to enhance air quality monitoring, offering actionable insights for urban planning and environmental protection. 
\end{abstract}

% keywords can be removed

\section{Introduction}

Air pollution ranks among the most critical environmental challenges of the 21st century, posing severe threats to human health, ecosystems, and the global climate \cite{manisalidis2020environmental, cohen2017estimates, pope2002lung}. The rapid growth of urbanization, industrialization, and reliance on fossil fuels has led to a sharp increase in atmospheric pollutants, including particulate matter (PM), nitrogen oxides (NOx), sulfur dioxide (SO2), carbon monoxide (CO), volatile organic compounds (VOCs), and ozone (O3). Prolonged exposure to these pollutants is linked to a wide range of health issues, such as respiratory diseases, cardiovascular disorders, and premature mortality. This underscores the urgent need for efficient and accurate systems to identify and monitor air pollutants. Among these pollutants, particulate matter has drawn particular attention due to its ability to transport other hazardous substances, penetrate deep into the respiratory tract, and even reach the bloodstream \cite{world2013health, myong2016health}.

Particulate matter, often referred to as aerosols, consists of solid or liquid particles suspended in the air. These particles are categorized by their diameter, such as PM2.5 (particles smaller than 2.5 $\mu$m) or PM10 (particles smaller than 10 $\mu$m). Composed of diverse chemical compounds, they primarily originate from combustion processes (e.g., residential heating, road transport), industrial activities, and agriculture \cite{pio2020source}. The formation and transformation of these particles in the atmosphere involve numerous complex processes \cite{pachauri2013characteristics, pant2013estimation}, making it challenging to forecast air quality, identify pollutant sources, and design effective mitigation strategies. Despite these complexities, identifying pollutant sources and assessing their composition is vital to implementing targeted health and environmental protection measures.

Traditional air quality monitoring relies on networks of physical sensors deployed across urban and industrial areas to measure pollutant concentrations. While these systems provide valuable data, they are often constrained by high deployment and maintenance costs, limited spatial coverage, and delayed reporting \cite{borrego2016assessment}. These limitations reduce their effectiveness in detecting sudden pollution events, identifying sources in real-time, and enabling proactive responses to mitigate air quality degradation.

Recent advances in micro-sensor technology offer a promising alternative for real-time air quality monitoring \cite{kumar2015rise, castell2017can}. These low-cost, portable devices can measure PM concentrations and particle size distributions with high spatial and temporal resolution, facilitating a more precise understanding of pollution patterns. However, while micro-sensors have significantly improved data collection capabilities, the challenge of pollutant source identification remains largely unaddressed.

Although research on real-time pollutant identification using advanced laboratory equipment has demonstrated the potential of machine learning (ML) in this field \cite{grant2018real}, to the best of our knowledge, no studies have yet employed ML techniques to identify pollutants in real time using data from micro-sensors. Previous research has primarily focused on predicting pollution levels \cite{kang2018air, ameer2019comparative, iskandaryan2020air}. The direct identification of pollutant types with micro-sensors, such as distinguishing between particulate matter originating from ash, sand, or candle emissions, represents a significant gap in the current literature. Addressing this challenge would enable a deeper understanding of pollution sources and provide actionable insights for urban planning and environmental protection.

In this work, we explore the innovative application of machine learning models \cite{bishop2006pattern, hastie2009elements} for real-time pollutant identification, leveraging PM data collected from micro-sensors. Specifically, we focus on the ability of these models to process PM data, adapt to dynamic environmental conditions, and identify pollutant sources with high precision. Our study introduces a novel classification framework capable of distinguishing four pollutant classes: Background Pollution, Ash, Sand, and Candle, in outdoor environment. By employing ML techniques, this research aims to pioneer the development of real-time pollutant identification systems, enhancing air quality management strategies and contributing to the development of sustainable cities and home.

The remainder of this paper is organized as follows. Section 2 presents the micro-sensor technology employed for pollutant identification and our methodology. Section 3 discusses the machine learning algorithms applied. Section 4 outlines the data collection. Section 5 details the experiments conducted and the results obtained. Finally, Section 6 concludes with insights and future directions.

\section{Materials and Methods}

\subsection{Optical micro-sensors}

Optical micro-sensors are widely employed for air quality monitoring, primarily functioning as particle counters \cite{karagulian2019review}. These devices utilize the scattering properties of light to estimate the size of airborne particles passing through the sensor. A controlled airflow is employed to guide particles into the sensor, where they are directed along a fluidic pathway, ensuring that individual particles intersect the laser beam. Upon interaction with the beam, each particle scatters light, which is then detected by one or more photodiodes. The particle mass concentrations for PM1, PM2.5, and PM10 are subsequently calculated using proprietary algorithms calibrated according to specific protocols \cite{wang2015laboratory, giordano2021low}. This calibration methodology may introduce biases when applied to real-world pollution conditions comprising heterogeneous particle compositions. To address this limitation, we based our analysis not on mass concentrations but on particle number within specific size ranges, described in Table \ref{tab:features}. This particle size distribution provides granular insight into pollution granulometry.

\begin{table}[H]
\caption{Particle sizes (in $\mu$m) description by channel (all the measurements are limited to 10.0$\mu$m)}
\label{tab:features}
\setlength{\tabcolsep}{4mm}
\small
\centering
\begin{tabular}{ccccc}
    \toprule
    \textbf{Chanel 1} & \textbf{Chanel 2} & \textbf{Chanel 3} & \textbf{Chanel 4} & \textbf{Chanel 5}   \\ 
    \midrule
     PM > 0.3 & PM > 0.5 & PM > 1.0 & PM > 2.5 &  PM > 5.0 \\ 
     \bottomrule
\end{tabular}
\end{table}
\vspace{-6pt}

In our study, we employed the optical particle sizer OEM sensor NextPS developed by TERA Sensor. To focus exclusively on its intrinsic performance, we opted to work directly with the standalone OEM sensor rather than one integrated into an air quality monitoring station. This approach allowed us to isolate and evaluate the sensor’s native capabilities, independent of external factors introduced by integration. 

\subsection{Methodology}

To disregard the intensity of the pollution phenomenon, we do not consider the absolute level of particulate matter (PM) directly. Instead, we focus on the ratios between the different levels of PM. Specifically, the inputs to our pollutant identification algorithms will not be $[PM>0.3, PM>0.5, PM>1.0, PM>2.5, PM > 5.0]$, but rather all the ratios:
\begin{align*}
    \left\{
    \begin{array}{l}
        \frac{PM>0.3}{PM>0.5} \\
        \vspace{0.02cm} \\
        \frac{PM>0.3}{PM>1.0} \\
        \vspace{0.02cm} \\
        \frac{PM>0.3}{PM>2.5} \\
        \vspace{0.02cm} \\
        ...
    \end{array}
    \right.
\end{align*}
This approach allows us to mitigate the impact of the intensity of the pollutant. Considering this method, the number of features changes, with 10 features instead of 5.

\section{Machine Learning Algorithms}

The task of pollutant identification is framed as a sequence modeling problem. For each $t > 0$, we observe a sequence of features $(\mathbf{x}_1, ..., \mathbf{x}_t)$ where $\mathbf{x}_t \in \mathbb{R}^{10}$ represents the feature vector at time $t$, derived from 5 particulate matter (PM) measurement channels and their 10 pairwise ratios. The objective is to classify the pollutant at time $t$, denoted $y_t$, with $y_t \in \Omega_Y = \{\text{Background Pollution, Ash, Sand, Candle}\}$. The classification at time $t$ may  leverage the feature sequence until $t$, $(\mathbf{x}_1, ..., \mathbf{x}_t)$.

To address this problem, we employ three widely recognized machine learning approaches suitable for sequential data analysis:
\begin{itemize}
    \item \textbf{XGBoost} \cite{chen2015xgboost}, a state-of-the-art ensemble learning algorithm leveraging decision trees \cite{breiman2017classification}; 
    \item \textbf{Long Short-Term Memory (LSTM)} networks \cite{hochreiter1997long}, a class of recurrent neural networks optimized for sequence dependency modeling \cite{lecun2015deep, goodfellow2016deep}; 
    \item \textbf{Hidden Markov Chain (HMC)} \cite{stratonovich1965conditional, baum1966statistical, cappe2006inference}, a probabilistic framework for modeling sequential data. 
\end{itemize}

In the following, we provide an in-depth discussion of these models, highlighting their theoretical foundations and application to the task.

\subsection{XGBoost}

XGBoost (eXtreme Gradient Boosting) is an efficient and scalable implementation of gradient boosting \cite{friedman2001greedy}, which is a machine learning technique used for supervised learning tasks such as classification and regression. XGBoost has gained significant attention due to its performance and speed. In segmentation tasks, XGBoost can be applied to predict the boundaries of segments within a sequence based on the features extracted from the data.

The key idea of XGBoost is to combine the predictions of several weak models, usually decision trees, to form a strong predictive model. The model iteratively refines predictions through boosting, where each new model corrects the errors of the previous models.

\subsection*{XGBoost Model Formulation}

The objective of XGBoost is to minimize the following loss function:

\[
L(\Theta) = \sum_{t=1}^{T} \ell(y_t, \hat{y}_t) + \sum_{k=1}^{K} \Omega(f_k)
\]

Where:
\begin{itemize}
    \item $\ell(y_t, \hat{y}_t)$: Loss function for the $t$-th data point. For our segmentation task, it is the cross-entropy.
    \item $\hat{y}_t$: Predicted value for the $t$-th data point.
    \item $\Omega(f_k)$: Regularization term to prevent overfitting. This term helps to control the complexity of the decision trees.
    \item $f_k$: The $k$-th weak learner (tree) in the ensemble.
\end{itemize}

The regularization term is typically given by:

\[
\Omega(f_k) = \gamma N + \frac{1}{2} \lambda \| \mathbf{w}_k \|^2
\]

Where:
\begin{itemize}
    \item $N$: The number of leaves in the tree.
    \item $\gamma$: A regularization parameter that controls the complexity of the tree.
    \item $\mathbf{w}_k$: The weights of the tree.
    \item $\lambda$: A regularization parameter for the weights.
\end{itemize}

\subsection*{Gradient Boosting Iteration}

At each iteration of the boosting process, XGBoost adds a new decision tree to the model. The update at iteration $n$ is computed as:

\[
\hat{y}_n = \hat{y}_{n-1} + \eta \cdot h_n
\]

Where:
\begin{itemize}
    \item $\eta$: The learning rate, which controls the contribution of each new model.
    \item $h_n$: The new weak model (tree) added at iteration $n$.
\end{itemize}

\subsection*{Segmentation Task}

In segmentation, XGBoost predicts the boundary or label of each segment in a sequence by learning from extracted features. Given a sequence $\{ \mathbf{x}_1, \mathbf{x}_2, \dots, \mathbf{x}_T \}$, the model outputs a sequence of predictions $\{ \hat{y}_1, \hat{y}_2, \dots, \hat{y}_T \}$ where each $\hat{y}_t$ indicates whether the corresponding data point $\mathbf{x}_t$ belongs to a specific segment or not.

\subsection{Long Short-Term Memory Networks}

Long Short-Term Memory networks are a type of recurrent neural network (RNN) designed to handle long-term dependencies. LSTMs are particularly suited for sequence data and have been successfully applied in segmentation tasks, where the goal is to divide a sequence into meaningful segments.

An LSTM unit consists of a cell, an input gate, a forget gate, and an output gate. These components enable the network to selectively remember or forget information.

\subsection*{LSTM Equations}

The LSTM network is governed by the following equations:

\begin{align*}
    \mathbf{f}_t &= \sigma(\mathbf{W}_f \mathbf{x}_t + \mathbf{U}_f \mathbf{h}_{t-1} + \mathbf{b}_f) \quad &\text{(Forget gate)} \\
    \mathbf{i}_t &= \sigma(\mathbf{W}_i \mathbf{x}_t + \mathbf{U}_i \mathbf{h}_{t-1} + \mathbf{b}_i) \quad &\text{(Input gate)} \\
    \mathbf{o}_t &= \sigma(\mathbf{W}_o \mathbf{x}_t + \mathbf{U}_o \mathbf{h}_{t-1} + \mathbf{b}_o) \quad &\text{(Output gate)} \\
    \tilde{\mathbf{c}}_t &= \tanh(\mathbf{W}_c \mathbf{x}_t + \mathbf{U}_c \mathbf{h}_{t-1} + \mathbf{b}_c) \quad &\text{(Cell candidate)} \\
    \mathbf{c}_t &= \mathbf{f}_t \odot \mathbf{c}_{t-1} + \mathbf{i}_t \odot \tilde{\mathbf{c}}_t \quad &\text{(Cell state update)} \\
    \mathbf{h}_t &= \mathbf{o}_t \odot \tanh(\mathbf{c}_t) \quad &\text{(Hidden state update)}
\end{align*}

Here:
\begin{itemize}
    \item $\mathbf{x}_t$: Input vector at time $t$.
    \item $\mathbf{h}_t$: Hidden state at time $t$.
    \item $\mathbf{c}_t$: Cell state at time $t$.
    \item $\mathbf{W}_f, \mathbf{W}_i, \mathbf{W}_o, \mathbf{W}_c$: Weight matrices for the input connections.
    \item $\mathbf{U}_f, \mathbf{U}_i, \mathbf{U}_o, \mathbf{U}_c$: Weight matrices for the recurrent connections.
    \item $\mathbf{b}_f, \mathbf{b}_i, \mathbf{b}_o, \mathbf{b}_c$: Bias terms.
\end{itemize}

For our experiments, we also add a linear layer followed by a softmax function. The hidden layer is composed of 50 neurons.

\subsection*{Segmentation Task}

In segmentation, the LSTM processes sequential input $\{\mathbf{x}_1, \mathbf{x}_2, \dots, \mathbf{x}_T\}$ and outputs a sequence $\{y_1, y_2, \dots, y_T\}$. By leveraging the memory mechanisms of LSTMs, the network can effectively learn dependencies across long sequences, making it a powerful tool for segmentation tasks.

The descent gradient algorithm with backpropagation \cite{lecun1988theoretical, lecun1989backpropagation} is used to learn the different parameters.

\subsection{The Hidden Markov Chain}

The Hidden Markov Chain is a powerful tool for supervised sequential segmentation, enabling the partitioning of sequential data into meaningful segments while accounting for temporal dependencies. In this context, the HMC leverages labeled training data to learn the underlying state-transition probabilities and observation emission probabilities, associating each segment with a distinct hidden state. By modeling both the sequential structure and the relationships between observations and states, the HMC provides a principled framework for segmenting sequences such as text, biological signals, or financial trends, ensuring that the segmentation aligns with the labeled patterns in the data.

\subsection*{Discriminative Forward Algorithm}

For segmentation purposes of the HMC, we apply the Discriminative Forward algorithm \cite{azeraf2020hidden, azeraf2023equivalence}, allowing to apply the HMC without modeling the observation's law. For each $t$, the estimation $\hat{y}_t$ given the previous observed sequence $\mathbf{x}_{1:t}$ is computed as follows:
\begin{align*}
    \hat{y}_t &= \arg \max_{\omega_i \in \Omega_Y} P(Y_t = \omega_i | X_{1:t} = x_{1:t}) \\
    &= \frac{\alpha_t(i)}{\sum\limits_{j=1}^{|\Omega_Y|} \alpha_t(j)}
\end{align*}
with:
\begin{itemize}
    \item For $t = 1$:
    \begin{align*}
        \alpha_1(i) = P(Y_1 = \omega_i | X_1 = \mathbf{x}_1);
    \end{align*}
    \item For $t > 1,$
    \begin{align*}
        \alpha_{t + 1}(i) = \frac{P(Y_{t + 1} = \omega_i | X_{t + 1} = x_{t + 1})}{P(Y_{t + 1} = \omega_i)} \sum\limits_{j=1}^{|\Omega_Y|} \alpha_t(j) P(Y_{t + 1} = \omega_i | Y_t = \omega_j).
    \end{align*}
\end{itemize}
We consider the homogeneous HMC, with the different law not depending on $t$. For each $(\omega_i, \omega_j)$, the laws $P(Y_{t + 1} = \omega_i)$ and $P(Y_{t + 1} = \omega_i | Y_t = \omega_j)$ are learned by counting frequencies, and $P(Y_{t + 1} = \omega_i | X_{t + 1} = \mathbf{x}_{t + 1})$ is modeled with a logistic regression and learned with gradient descent algorithm and backpropagation, as the LSTM.

\section{Data Collection}

There is currently no publicly available dataset for real-time pollutant identification using a micro-sensor. As a result, we created our own dataset through the experimental procedures detailed in this section.

We identified four pollutant cases for this study, chosen based on their prevalence and potential for environmental impact. The pollutants studied include:
\begin{itemize}
    \item Sand
    \item Ash
    \item Candles
    \item Background Pollution
\end{itemize}

\begin{figure}[t]
    \centering
    \includegraphics[scale=1]{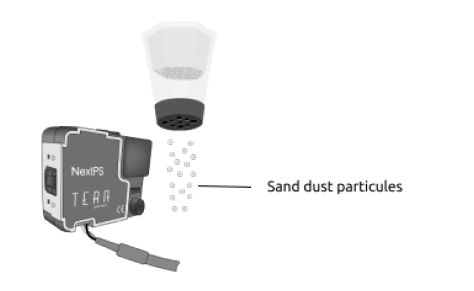}
    \captionof{figure}{Example of data collection with sand as pollutant}
    \label{sensor_collection}
\end{figure}

The data collection process was conducted using an optical micro-sensor NextPS located outdoors in Aix-en-Provence, southern France. The sensor was exposed to both ambient environmental conditions and controlled emissions of specific pollutants, such as sand, ash, and candles. Controlled experiments were designed to simulate diverse real-world scenarios, varying the duration and intensity of pollutant exposure. The process is illustrated in Figure \ref{sensor_collection}.

Data were collected across the five sensor channels, capturing detailed particulate matter (PM) concentrations. These measurements formed the basis of the dataset used to train and evaluate the machine learning models described in Section 3.

The dataset consisted of samples collected under the following conditions:
\begin{itemize}
    \item Background Pollution: Four sessions -- three lasting 180 seconds and one lasting 60 seconds.
    \item Sand: Two sessions, each lasting 180 seconds.
    \item Ash: Three sessions -- lasting 60, 120, and 180 seconds, respectively.
    \item Candles: One session of 180 seconds.
\end{itemize}

\begin{figure}[t]
    \centering
    \begin{minipage}{0.45\textwidth}
        \centering
        \includegraphics[width=\textwidth]{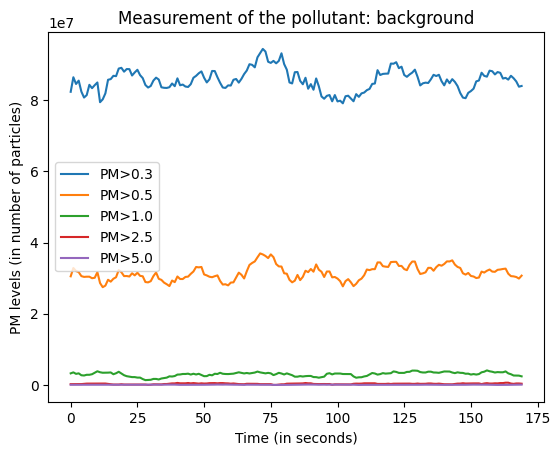}
    \end{minipage}
    \hfill
    \begin{minipage}{0.45\textwidth}
        \centering
        \includegraphics[width=\textwidth]{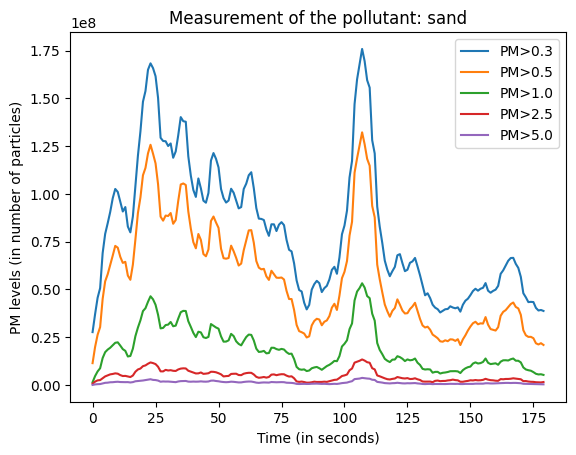}
    \end{minipage}
    \vskip\baselineskip
    \begin{minipage}{0.45\textwidth}
        \centering
        \includegraphics[width=\textwidth]{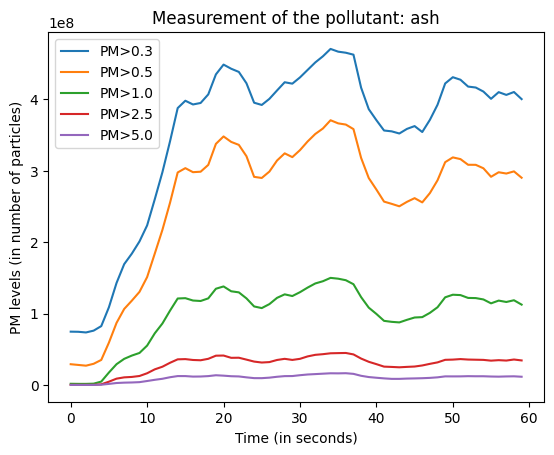}
    \end{minipage}%
    \hfill
    \begin{minipage}{0.45\textwidth}
        \centering
        \includegraphics[width=\textwidth]{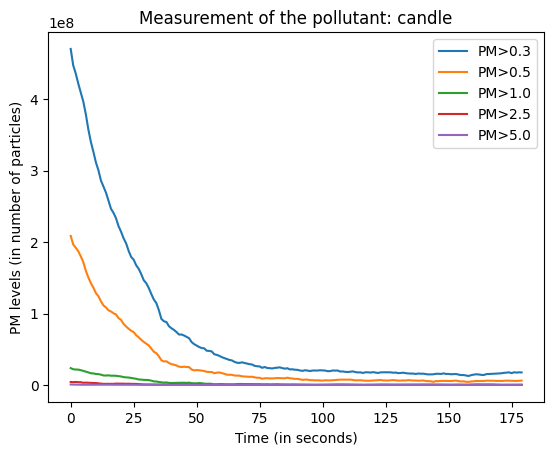}
    \end{minipage}
    \caption{Representation of the channels for each scenario - different scales per scenario are used for clarity.}
    \label{data_representation}
\end{figure}

\paragraph{}
Representative visualizations of the data for each pollutant are shown in Figure \ref{data_representation}.

\section{Model training and Results}

We are facing a supervised multi-label classification task. The dataset was split into training (70\%) and testing (30\%) subsets without shuffling to preserve temporal dependencies. This division resulted in 1,050 data points for the training set and 450 data points for the testing set. The training set was used to optimize the parameters of the XGBoost, LSTM, and Hidden Markov Chain (HMC) models, while the testing set was reserved for evaluating performance on unseen data.

All models were trained to minimize the Cross-Entropy loss function. More specifically:
\begin{itemize}
    \item The XGBoost model was implemented using the xgboost Python library and trained for 100 boosting iterations.
    \item The LSTM model was implemented with the PyTorch \cite{paszke2019pytorch} library and trained for 1,000 iterations on the entire training set using the Adam optimizer \cite{kingma2014adam}, with a learning rate of 0.001.
    \item About the HMC, the transition probabilities were estimated by counting frequencies in the training set. Specifically, for any pair of states $\omega_i$ and $\omega_j$, the probabilities were computed as:
    \begin{align*}
        \hat{P}(Y_{t + 1} = \omega_i) &= \frac{N(\omega_i)}{\sum\limits_k N(\omega_k)}, \\
        \hat{P}(Y_{t + 1} = \omega_i | Y_t = \omega_j) &= \frac{N(\omega_j, \omega_i)}{\sum\limits_k N(\omega_j, \omega_k)}
    \end{align*}
    where $N(\omega_i)$ denotes the number of occurrences of $\omega_i$ in the training set, and $N(\omega_j, \omega_i)$ represents the number of times $\omega_j$ is followed by $\omega_i$. The probabilities $P(Y_{t + 1} = \omega_i | X_{t + 1} = \mathbf{x}_{t + 1})$ were learned via gradient descent using the PyTorch library. Training was conducted for 5,000 iterations with the Adam optimizer, employing a learning rate of 0.01.
\end{itemize}
The parameters of the training have been determined thanks to a validation step.

Accuracy, defined as the percentage of correctly identified pollutants, served as the primary evaluation metric. We also compute the F1 score for each case for each pollutant. The F1-score is a measure that considers both precision and recall. It is defined as the harmonic mean of precision and recall:
\begin{equation*}
F_1 = 2 \times \frac{\text{Precision} \times \text{Recall}}{\text{Precision} + \text{Recall}}
\end{equation*}
where precision and recall are given by:
\begin{equation*}
\text{Precision} = \frac{TP}{TP + FP} \;\;\;\;\;\; \text{Recall} = \frac{TP}{TP + FN}
\end{equation*}
where \( TP \), \( FP \), and \( FN \) denote the number of true positives, false positives, and false negatives, respectively. The F1-score balances the trade-off between precision and recall, making it particularly useful for imbalanced classification problems.

The results for accuracy are summarized in Table \ref{acc_table}. They are in Tables \ref{f1_table} for the F1 scores. In this table, we also compute the weighted F1 scores of each model.

\begin{table}[t]
\caption{Accuracy of XGBoost, LSTM, and HMC, for real-time pollutant identification}
\centering
\begin{tabular}{c|c|c}
    \toprule
    \textbf{XGBoost} & \textbf{LSTM} & \textbf{HMC} \\ 
    \midrule
     81.56\% & 77.78\% & \textbf{82.44\%}  \\ 
     \bottomrule
\end{tabular}
\label{acc_table}
\end{table}

\begin{table}[t]
\caption{F1 Scores of the XGBoost, LSTM, and HMC, for each pollutant for real-time pollutant identification}
\centering
\begin{tabular}{c|c|c|c}
    \toprule
    {} & \textbf{XGBoost} & \textbf{LSTM} & \textbf{HMC} \\
    \midrule
    \textbf{Ash} & 94.64 & \textbf{98.63} & 79.80 \\ 
    \textbf{Background} & 81.89 & 76.40 & \textbf{88.27} \\
    \textbf{Candle} & \textbf{72.38} & 0.00 & 64.46 \\ 
    \textbf{Sand} & 69.05 & 79.07 & \textbf{85.71} \\
    \midrule
    \textbf{Weighted F1 Score} & 81.73 & 73.21 & \textbf{82.76} \\
    \bottomrule
\end{tabular}
\label{f1_table}
\end{table}

\begin{figure}[H]
    \centering
    \includegraphics[scale=0.5]{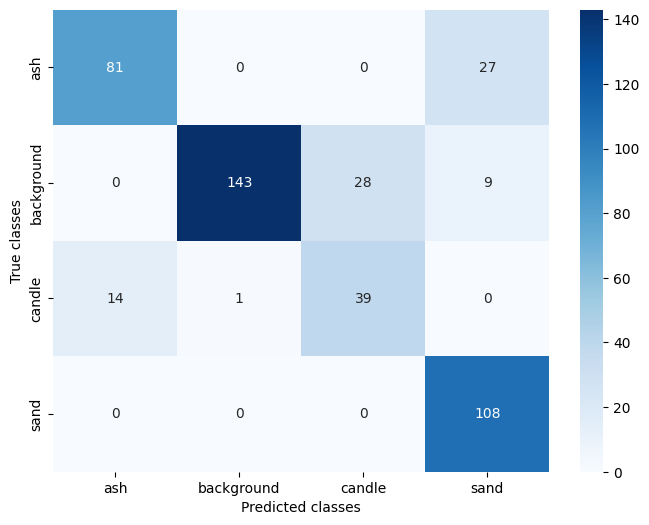}
    \captionof{figure}{Confusion Matrix of the HMC model}
    \label{confusion_matrix}
\end{figure}

Among the three models, HMC demonstrated the highest accuracy and weighted F1-Score, outperforming both the sequential LSTM model and the XGBoost. It is also the best model to identify background and sand. The confusion matrix for HMC, in Figure \ref{confusion_matrix}, provides additional insight into its classification performance.

Using this model, for instance with ash as the pollutant, the model correctly identifies ash in 81 cases over the 108. However, it mistakenly predicts sand 27 times whereas ash is the true pollutant, and incorrectly predicts ash in 14 instances when candle is actually present.

\section{Discussion}

The results highlight HMC's superior performance in real-time pollutant identification, achieving a accuracy of 82.44\%. XGBoost, while slightly less precise, also demonstrated promising results. Conversely, the LSTM model underperformed, despite its inherent suitability for sequential data modeling, due to its inability do identify the candle.

The confusion matrix analysis for HMC revealed strong identification rates for ash. However, challenges were noted in distinguishing between certain pollutants, particularly background and candles.

We also observe that the models achieve good results even despite a small dataset to train them. 

It is now important to test these models with a much larger dataset to assess the impact on model accuracy. Some models as LSTM have shown better performance with a larger dataset in time classification exercise \cite{farahani2025time}. Whereas, the diversity of pollution events with various characteristics and type could bring confusion in models results. Finally, the models have not be tested on mixed pollution events which can occur in real environment due to mixing of sources in the atmosphere.

\section{Conclusion}

This study successfully demonstrated the potential of optical micro-sensors for real-time pollutant identification in outdoor settings. Despite relying on a low-cost micro-sensor and limited data collection durations, the machine learning models achieved commendable accuracy.

HMC emerged as the most effective model, achieving an accuracy of 82.44\%, showcasing its robustness for this application. These findings underscore the potential of this model for practical deployments in pollutant monitoring.

Future work could extend this research by collecting data in specialized environments, such as construction sites or subway stations, to evaluate the models' adaptability. Additionally, exploring advanced extensions of Hidden Markov Chains \cite{pieczynski2003pairwise, azeraf2021cist, azeraf2021introducing} could offer valuable insights, particularly given their promising performance on limited datasets.

%%%%%%%%%%%%%%%%%%%%%%%%%%%%%%%%%%%%%%%%%%
\paragraph{Author Contribution}{Experimental work: E.A., A.W. and E.B.; methodology: E.A., A.W. and E.B.; machine learning: E.A. and A.W.; code implementation: E.A.; result analysis and discussion: E.A., A.W. and E.B.; writing: E.A. (Section 1, 2.2, 3, 4, 5, 6, 7) and E.B. (Section 2.2); reviewing: L.L. and S.M. All authors have read and agreed to the published version of the manuscript.}

\section*{Appendix}

The confusion matrices of the XGBoost and the LSTM models are provided below.

\begin{figure}[H]
    \begin{minipage}{0.45\textwidth}
        \centering
        \includegraphics[width=\textwidth]{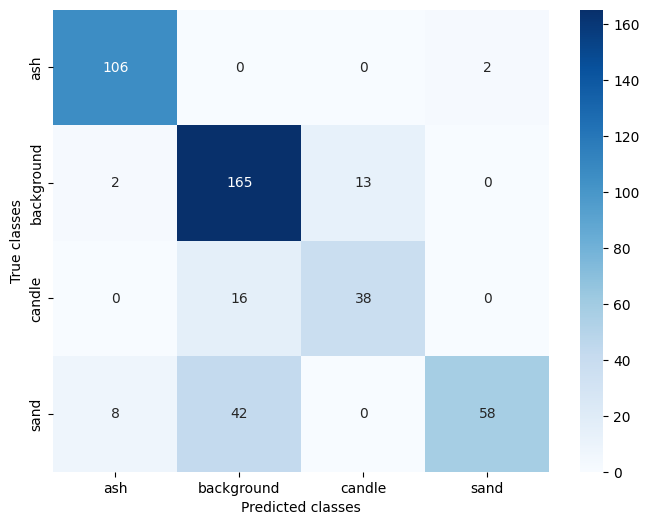}
    \end{minipage}
    \hfill
    \begin{minipage}{0.45\textwidth}
        \centering
        \includegraphics[width=\textwidth]{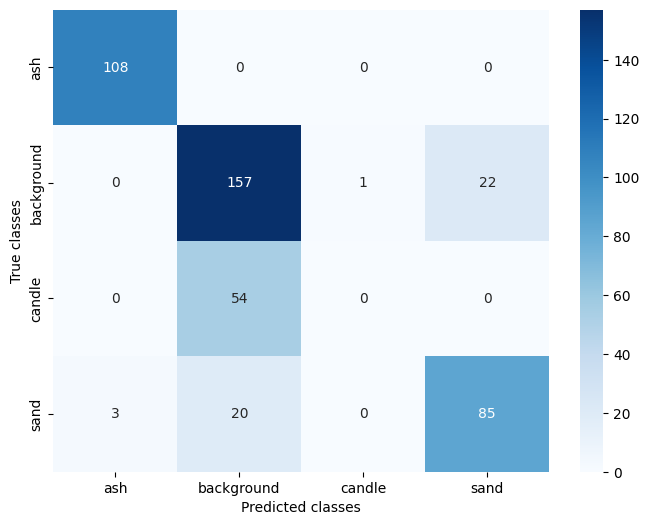}
    \end{minipage}
    \captionof{figure}{Confusion matrices of the XGBoost and the LSTM models, respectively}
\end{figure}

\bibliographystyle{unsrt}  
\bibliography{references}  %%% Remove comment to use the external .bib file (using bibtex).

\end{document}